\documentclass[10pt,twocolumn,letterpaper]{article}

\usepackage{cvpr}
\usepackage{times}
\usepackage{epsfig}
\usepackage{graphicx}
\usepackage{amsmath}
\usepackage{amssymb}
\usepackage{helvet} 
\usepackage{courier}  
\usepackage[hyphens]{url}  
\usepackage{cite}
\usepackage{caption}
\usepackage{subcaption}
\usepackage[acronym]{glossaries}
\usepackage{booktabs}
\usepackage{amsfonts}

\cvprfinalcopy 

\def\cvprPaperID{****} 

\newacronym{gan}{GAN}{Generative Adversarial Network}
\newacronym{vae}{VAE}{Variational  Autoencoder}
\newacronym{caae}{CAAE}{Conditional Adversarial Autoencoder}
\newacronym{fiw}{FIW}{Families In the Wild}
\newacronym{dna}{DNA}{Deoxyribonucleic acid}

\begin{document}

\title{What Will Your Child Look Like? DNA-Net: Age and Gender Aware Kin Face Synthesizer}

\author{Pengyu Gao\\
Southeast University\\
{\tt\small pi\_1412@163.com}
\and
Siyu Xia\\
Southeast University\\
{\tt\small xia081@gmail.com}
\and
Joseph Robinson\\
Northeastern University\\
{\tt\small robinson.jo@husky.neu.edu}
\and
Junkang Zhang\\
University of California San Diego\\
{\tt\small juz007@eng.ucsd.edu}
\and
Chao Xia\\
ShangHai Jiao Tong University\\
{\tt\small xiabc612@gmail.com}
\and
Ming Shao\\
University of Massachusetts Dartmouth\\
{\tt\small mshao@umassd.edu}
\and
YUN FU\\
Northeastern University\\
{\tt\small yunfu@ece.neu.edu}
} 

\maketitle

\begin{abstract}
   Visual kinship recognition aims to identify blood relatives from facial images. Its practical application-- like in law-enforcement, video surveillance, automatic family album management, and more-- has motivated many researchers to put forth effort on the topic as of recent. In this paper, we focus on a new view of visual kinship technology: kin-based face generation. Specifically, we propose a two-stage kin-face generation model to predict the appearance of a child given a pair of parents. The first stage includes a deep generative adversarial autoencoder conditioned on ages and genders to map between facial appearance and high-level features. The second stage is our proposed DNA-Net, which serves as a transformation between the deep and genetic features based on a random selection process to fuse genes of a parent pair to form the genes of a child. We demonstrate the effectiveness of the proposed method quantitatively and qualitatively: quantitatively, pre-trained models and human subjects perform kinship verification on the generated images of children; qualitatively, we show photo-realistic face images of children that closely resemble the given pair of parents. In the end, experiments validate that the proposed model synthesizes convincing kin-faces using both subjective and objective standards.
\end{abstract}

\section{Introduction}

The goal of automatic kinship recognition is to determine whether or not people are related, and furthermore if so, the type of relationship shared. In the visual domain, faces are typically used as the cue to determine kinship. This technology can be applied to mine social relationship~\cite{Zhang2017Family}, build a family tree~\cite{Chao2018Graph}, aid criminal investigations, do nature-based studies~\cite{crouch2018genetics}, and more. From this, kinship recognition has gained the interest of vast researchers nowadays.

In this work, we tackle a different task than is traditionally addressed in of kinship recognition, \ie kin-face generation. Our aim to predict the appearance of a child from a pair of parents conditioned on high-level features (\ie age and gender), which provides control over the desired characteristics.

\begin{figure}[t!]
    \centerline{
        \includegraphics[width=0.9\linewidth]{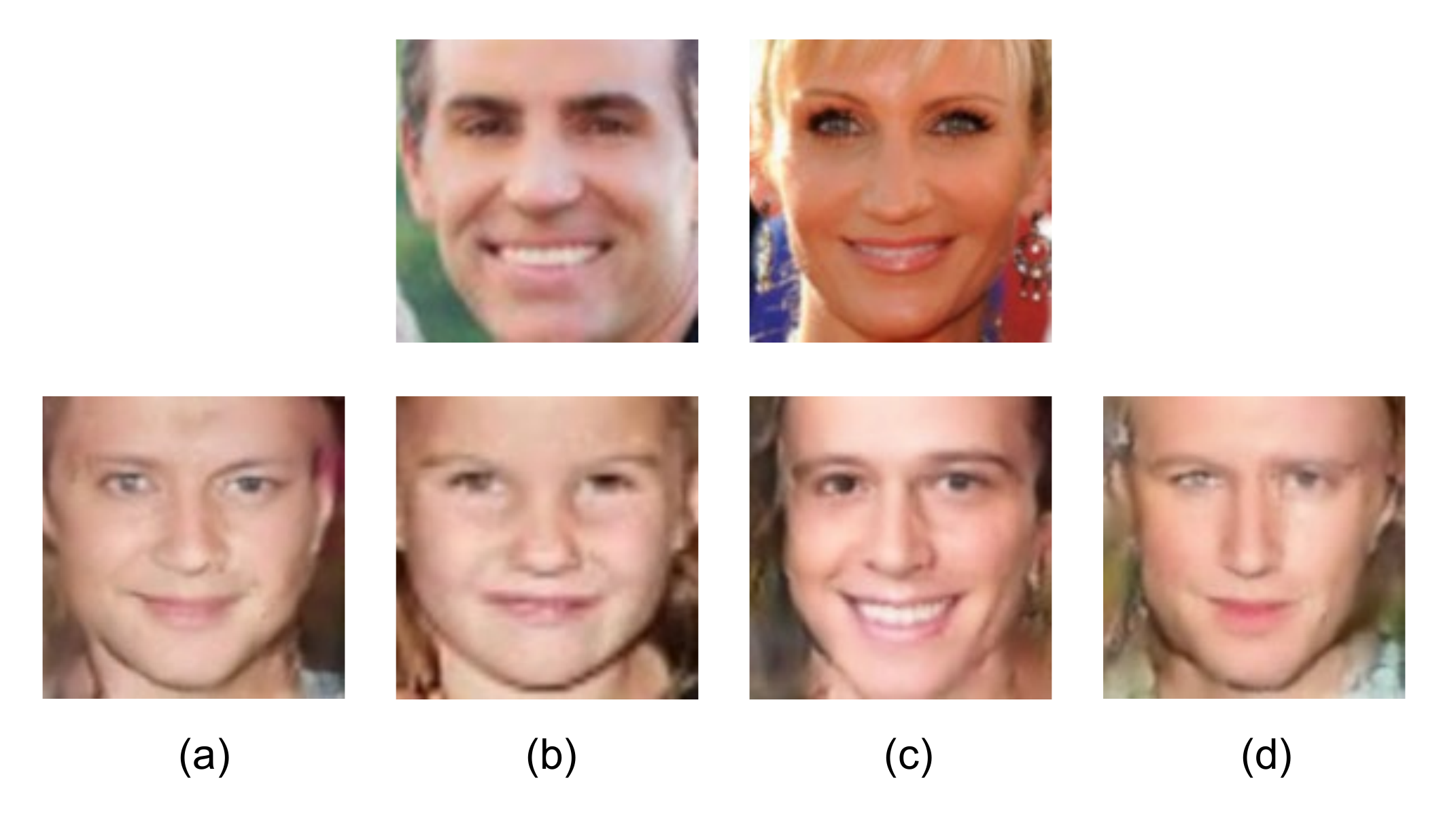}}
    \caption{From faces of parents (top row), which face resembles their child the most (bottom row)? Three of the faces are generated, while one is real. Can you guess which one?}
    \label{fig:kin}
\end{figure}

The biological mechanisms that drive the visual resemblance of parents and their children inspired our efforts, and thus ability, to automatically understand kin-faces~\cite{alvergne2007differential, debruine2008social}. Daly and Wilson~\cite{daly1982whom} hypothesized that face similarity is sufficient evidence for kinship. Naini and Moss~\cite{naini2004three} claimed to have cracked the code for finding the most critical genetic features, which they quantified as ``relatedness''. More recently, researchers generated heritability maps that link facial landmarks to specific phenotypes of twins~\cite{tsagkrasoulis2017heritability}]. The generated maps were from high-resolution faces (\ie 4,096 landmarks) of 954 twins captured by expensive 3D cameras, which the authors identified genetic correspondents in the face variations of twins.

Typically, two directions are followed to recognize kin-faces: hand-crafted features and metric-based learning. Nowadays, deep models, especially Convolutional Neural Network (CNN), have shown promising discriminative power when used to encode faces for kinship recognition, pushing the state-of-the-art in the verification (\ie one-to-one) task~\cite{zhang2015kinship, lu2017discriminative}.

Out of the many recent works in automatic kinship, only a few have attempted the kinship generation problem. Ertugrul~\etal~\cite{ertugrul2017dynamic} focused on generating the facial dynamics of a child (\eg smile) from a video of a parent showing different facial expressions. Ozkan~\etal~\cite{ozkan2018kinshipgan} generated a child's face, given a parent via adversarial training with constraints on the gender class and cycle consistency. Note, existing approaches that generate kin-faces, although unique in their ways, share a common flaw-- only a single parent used to predict faces of children. These methods are unable to incorporate information from a pair of parents-- the results are ineffective when compared to true child. Furthermore, they do not properly mimic nature (\ie it takes two to reproduce).

In summary, the process of inheritance can be generalized in two main steps: (1) the local traits and global shape of the face are mostly determined by genes controlling the production of proteins at the micro-level and (2) genes of an offspring are inherited from one parent or the other by a random selection and combination process. Thus, children are not identical to a single parent but tend to resemble both parents in various ways. The practical significance of predicting the appearance of a child from a parent pair should be acknowledged, and the existing methods based on single inputs should be christened limited and unrealistic.

To incorporate the concepts of genetics into the kinship generation problem, we utilize an encoder-decoder structure~\cite{makhzani2015adversarial, larsen2015autoencoding} to mimic the process of inheritance in facial appearance by transforming genes from parents-to-child.
Previously, the encoder-to-decoder structure has been incorporated into \gls{gan} \cite{goodfellow2014generative} and \gls{vae}~\cite{kingma2013auto} to generate photo-realistic faces~\cite{radford2015unsupervised}, where mappings between facial images and high-level personal features were established.
Similarly, in our kinship generation task, the facial traits of parents (\ie an image pair) are translated into genes by the encoder. Then, the child's genes can be generated by simulating the random selection and combination process on the gene-encodings of the parents. Finally, the face of the child is generated by decoding the genes.



We propose a kinship generation model with a two-step learning procedure inspired by the genetic process. Step one: a deep generative \gls{caae}~\cite{zhang2017age} is trained on a large-scale face dataset to learn to map facial appearance to high-level features with knowledge of age and gender. Step two: a novel DNA-Net, trained on a smaller kinship dataset, transforms high-level features to genes, \ie translates genes of a parent pair to a child. Figure~\ref{fig:kin} depicts the inputs and outputs of the proposed model. Can you determine which are the real children (bottom row) of the parents (top row)?


There are two main contributions in this paper.
\begin{enumerate}
    \item  We introduce DNA-Net to transfer features from parents to child by simulating the genetic process, while combining it with the \gls{caae} model to realize child facial image generation from the images of parents.
    \item We are able to generate multiple siblings by manipulating the gene codes in DNA-Net, which allow for changes to be made to the generated child in both age and gender.
\end{enumerate}
\textit{Beyond these contributions, we plan to promote our methodology with broader impacts through crowd-sourcing: Given enough data, our model will be able to reveal the mechanism and hidden factors of gene combination from parents, which is less random and more governed by natural laws.}

\section{RELATED WORK}
\label{sec:format}

\subsection{Kinship Verification}
The task of kinship verification is to determine whether a face pair is related (\ie KIN or NON-KIN). Evaluations are typically done separately for different relationship, like parent-child, siblings, and sometimes grandparent-grandchild. Research in both psychology and computer vision revealed that different kin relations render different familial features, which motivated researchers to model different relationship types independently. Existing methods for the kinship verification can generally be split into either metric learning based \cite{lu2014neighborhood, yan2014discriminative} or feature based methods \cite{ertugrul2017dynamic}. In metric learning methods, either a distance measure or feature transformation is learned to reduce distances between kin pairs and push away non-kin pairs. Feature based methods use hand-crafted features or learn more discriminative representations.

Recently, deep neural networks have achieved state-of-the-art in kinship verification. \cite{dehghan2014look} proposed a method to discover the optimal features and metrics that relate a parent to offspring via gated autoencoders. \cite{zhang2015kinship} utilized CNNs as a feature extractor for kinship verification. ~\cite{wang2018photo} integrated the triple ranking loss into CNN model to learn more discriminative representations.

Some methods incorporate deep metric learning for better performance. \cite{wang2017kinship} proposed a denoising auto-encoder based on marginalized metric learning to preserve the structure of data and simultaneously endow the discriminative information into the learned features. ~\cite{lu2017discriminative} developed a discriminative deep multi-metric learning method to jointly learn multiple neural networks to better use the commonality of multiple
feature descriptors. See past challenges for various other methods and task specific information~\cite{lu2015fg}.

\begin{figure}[t!]
    \centerline{\includegraphics[width=8.5cm]{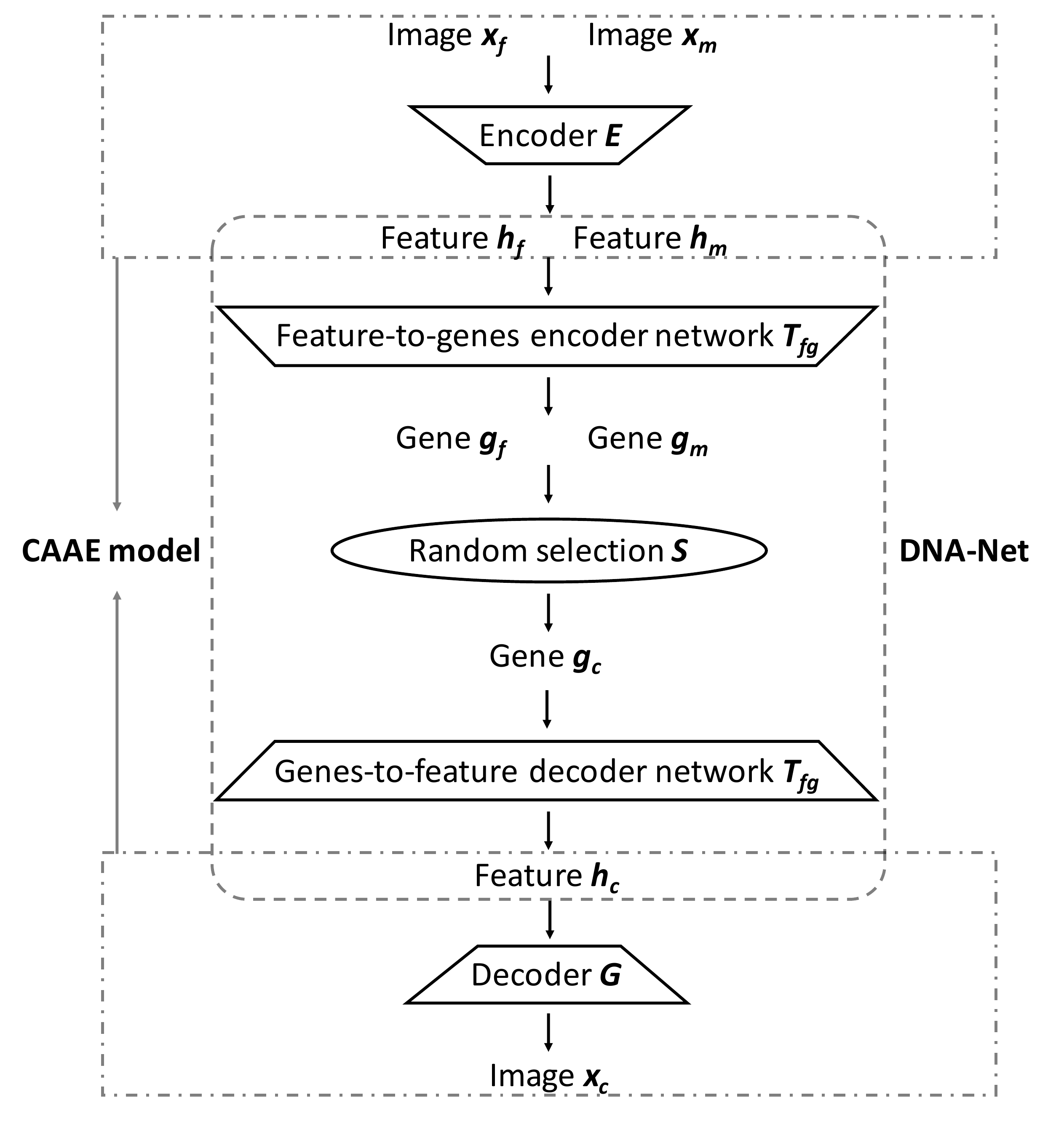}}
    \caption{Flowchart of genetic model. Note that the variables used are consistent with that in Eq.~\ref{eq:face1}-\ref{eq:dnaminmax} and Figure~\ref{fig:dnanet}.}
\label{fig:geneticmodel}
\end{figure}

\begin{figure*}[t!]
    \centerline{
        \includegraphics[width=\linewidth]{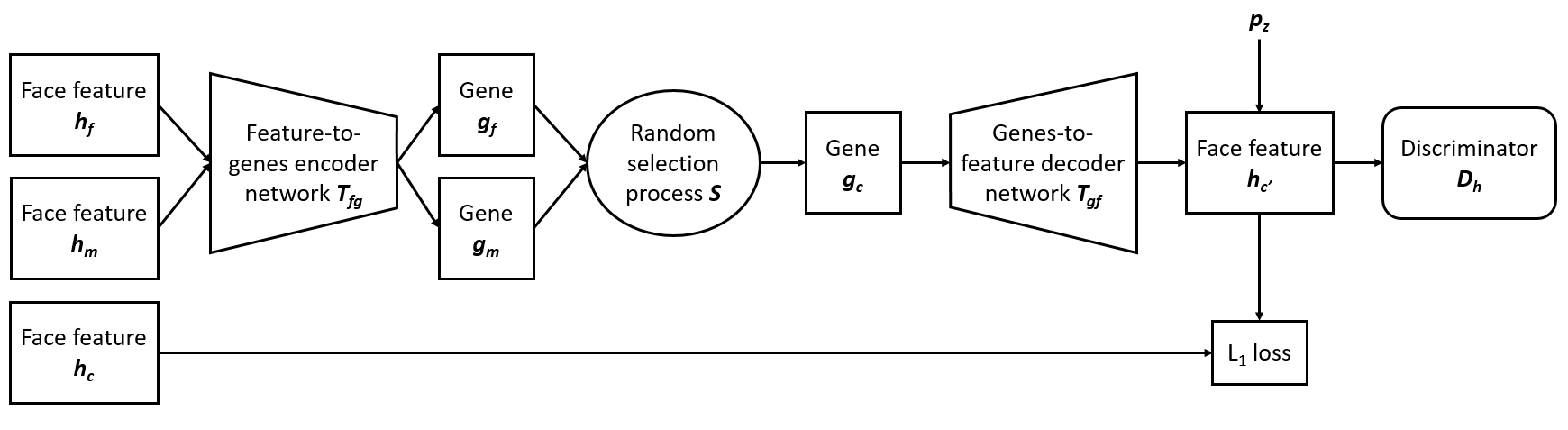}
        }
    \caption{Architecture of DNA-Net. The encoder network $T_{fg}$ maps the extracted face feature $h$ to gene $g$. Random selection process $S(\cdot)$ transfer genes from parents to child. Decoder network $T_{gf}$ maps gene $g$ to feature $h$. The discriminator $D_h$ imposes the uniform distribution on $h$ and $p_z$ is a prior distribution. The network updated based on the $L_1$ loss between the input face feature $h_c$ and generated face feature $h_{c'}$ of child. Note that $f$, $m$, $c$ in the figure means father, mother, child respectively.}
    \label{fig:dnanet}
\end{figure*}

\subsection{Deep Generative Models}
\gls{vae} and \gls{gan} are two of the most renowned image generation models. Both methods can generate images from latent codes that follow certain prior distributions. In recent years, multiple variants of these two have emerged. Some of them adopt an encoder-decoder structure that can also map images into latent codes which can be considered as features. In \cite{isola2017image}, Isola proposed pix2pix, an image-to-image translation method based on conditional \gls{gan} (cGAN) \cite{mirza2014conditional}. Pix2pix can be seen as learning two mappings, image to features and features to image. Then came inverted conditional \gls{gan} (IcGAN) \cite{perarnau2016invertible}, a two-step image-to-image translation method which focuses on face attributes editing, like transforming smiling face to non-smiling face. In IcGAN, an additional encoder is trained to map a image into latent codes/features and conditional representation after a cGan was trained first. After the training of additional encoder, face attributes can be changed by manipulation of latent codes. In \cite{wang2017tag}, a tag mapping net was proposed which maps tags (labels) of image to features which are encoded from image, making it possible to adjust the attributes of generated image by adjusting the tag. \cite{xiao2018elegant} proposed an image-to-image translation model which focuses on face attributes editing and can deal with multiple face attributes simultaneously. These works give us inspiration that the mapping between face image and face features can be learned in deep generative models~\cite{isola2017image,perarnau2016invertible}, even mapping between features and features can be learned (tags can be seen as kind of features)~\cite{wang2017tag}, and image content can be manipulated with latent codes~\cite{perarnau2016invertible}.

A special variant of \gls{vae} and \gls{gan} is the combination of the two, with \gls{vae}/\gls{gan}~\cite{larsen2015autoencoding} and AAE~\cite{makhzani2015adversarial} being amongst the most popular. When used together, these models inherit the ability of inference from \gls{vae} and the tendancy to generate sharp pictures of \gls{gan}. Also, \gls{vae}/\gls{gan} and AAE have encoder-decoder structures.

\section{APPROACH}
\label{sec:pagestyle}
First, we use neural network terminology to model the genetic process. Then, a \gls{caae} model adapts to establish two-way mappings between facial images and face features. Finally, our DNA-Net establishes two-way mappings between face features and genes, \ie analogous to inheritance.

\subsection{Genetic Model}

Research in genetics revealed that multiple genes could contribute to a single facial trait, for instance, 16 genes were found to effect eye color~\cite{white2011genotype}. From this, translating from genes to face appearance is modeled as
\begin{equation}
\begin{aligned}
\label{eq:face1}
x_k = F(g_{k1},g_{k2},...,g_{kn}),  \quad k \in  \left\{f,m,c\right\}
\end{aligned}
\end{equation}
where $f$, $m$, $c$ stand for father, mother, and child, respectively, $x_k$ is the appearance of $k$, and $g_{ki}$ denotes the genes responsible for the facial features. $F(\cdot)$ produces a face based on gene(s). As shown in nature, a child genetically inherits genes from both parents via random selection. This random selection can be expressed as follows:
\begin{equation}
\begin{aligned}
\label{eq:face2}
x_c &= F(g_{c1},g_{c2},...,g_{cn})\\
&= F(S(g_{f1},g_{m1}),S(g_{f2},g_{m2}),...,S(g_{fn},g_{mn}))),
\end{aligned}
\end{equation}
where $S(\cdot)$ simulates the process of obtaining the gene of a child $g_{ci}$ through a random selection over the corresponding genes of the two parents, which is thus defined as
\begin{equation}
\begin{aligned}
\label{eq:gene1}
g_{ci} &= S(g_{fi},g_{mi})\\
&= r_i \cdot g_{fi} + (1-r_i) \cdot g_{mi} ,  \quad  r_i \in \left\{0,1\right\},
\end{aligned}
\end{equation}
where $r_i$ is a value randomly assigned.

To incorporate the process of Eq. (\ref{eq:face1})-(\ref{eq:gene1}) into our generative model, we design a genetic model that generates a face of a child from faces of a pair of parents. This model contains three main stages. Figure~\ref{fig:geneticmodel} depicts this genetic model.

\textbf{First stage}. Genes of parents are predicted from their appearances. Specifically, we encode faces to represent $x_k$, $k\in\{f,m\}$ and generate personal facial features $h_k$ with encoder $E$ through
\begin{equation}
h_{k} = E(x_{k}), \quad k \in  \left\{f,m\right\}.
\label{eq:feat1}
\end{equation}
Feature vectors $h_k$ will then be translated to gene vectors $g_k$ by another gene encoder $T_{fg}$ as
\begin{equation}
g_{k} = [g_{k1}, g_{k2}, ...] = T_{fg}(h_{k}), \quad k \in  \left\{f,m\right\}.
\label{eq:gene2}
\end{equation}

\textbf{Second stage}. We derive the gene vector of the child $g_c$ from the genes of the parents via a random selection process over corresponding gene elements. This can be expressed as
\begin{equation}
g_{c} = [g_{c1}, g_{c2}, ...] = [S(g_{f1},g_{m1}), S(g_{f2},g_{m2}), ...].
\label{eq:gene3}
\end{equation}

\textbf{Third stage}. We predict the facial appearance of the child $x_c$ from genes $g_c$ output from two decoders. Specifically, the personal facial feature $h_c$ is decoded from gene $g_c$ by a gene decoder $T_{gf}$ as
\begin{equation}
h_{c} = T_{gf}(g_{c}).
\label{eq:feat2}
\end{equation}
Then, the facial image is generated by another decoder $G$:
\begin{equation}
x_{c} = G(h_{c}).
\label{eq:face3}
\end{equation}
Eq. (\ref{eq:face1}) can be represented as $x_k=F(g_k)=G(T_{gf}(g_k))$.

We use \gls{caae}~\cite{zhang2017age} to train the image-feature encoder $E$ and decoder $G$. Then, a novel neural network dubbed DNA-Net was designed to model the mappings between extracted features and genes via $T_{fg}$ and $T_{gf}$ as well as the random selection process $S(\cdot,\cdot)$. One reason to use separate networks $T_{gf}(\cdot)$ and $G(\cdot)$, instead of a single network for $F(\cdot)$ (and vice versa) is that, the limited amount of data labeled for kinship recognition is less suited to support training of a single larger network that directly maps between images and genes (\ie prone to overfitting). Instead, we choose to train $E(\cdot)$ and $G(\cdot)$ in the \gls{caae} on a large-scale face dataset, and then train the smaller DNA-Net on the smaller kinship dataset. Besides, we want encoder $E$ and decoder $G$ to capture age and gender information, opposed to DNA-Net, as most genes are age-invariant.

\subsection{Image-Feature Mapping via \gls{caae}}



%

Next, we discuss the details of \gls{caae}~\cite{zhang2017age}. The input and output of \gls{caae} net are $128 \times 128$ RGB facial images
$x \in R^{128 \times 128 \times 3}$. On the one hand, the encoder $E(\cdot)$ preserves the high-level personal features of the input face $x$ in a feature vector $h=E(x) \in R^n$. On the other hand, the decoder $G$ generates a face image $\hat{x}=G(h,l)$ that is conditioned on a certain age and gender. Note that $l$ is a one-hot vector encoding age and gender labels. In the end, the input and output faces aim to be as similar as possible:
\begin{equation}
\mathop{\min}\limits_{E,G}L(x,G(E(x),l)),
\end{equation}
where $L(\cdot , \cdot)$ denotes euclidean distance.

\begin{figure}[t!]
    \centerline{
        \includegraphics[width=8.5cm]{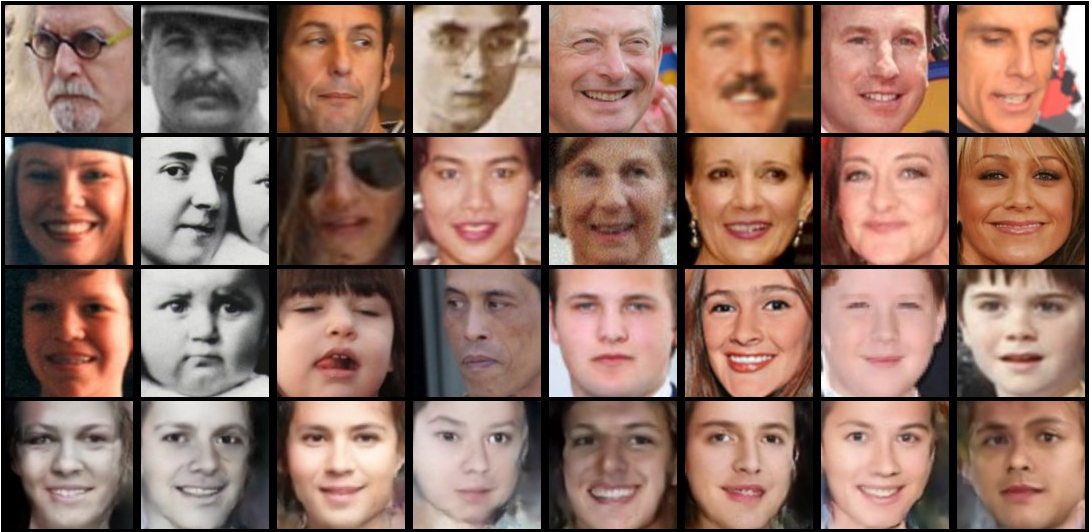}
        }
    \caption{Samples results. Each column corresponds to a family, with faces of fathers on first row, mothers on second, real children on third, and generated children on bottom.}
    \label{fig:samles results}
\end{figure}

Additionally, two discriminator networks, $D_{z}$ and $D_{img}$, are placed after $E$ and $G$, respectively, for the purpose of adversarial training. $D_{z}$ regularizes the feature vector $h$ to be uniformly distributed to smooth the age transformation.
We denote the distribution of the training data as $p_{data}(x)$, while the distribution of feature $h$ is $q(h|x)$. Also, $p(z)$ is assumed to be a prior distribution, and $z^{*} \sim p(z)$ denotes the random sampling process from $p(z)$. A min-max objective function can be used to train $E$ and $D_{z}$ as
\begin{equation}
\begin{aligned}
\label{eq:dz}
\mathop{\min}\limits_{E} \mathop{\max}\limits_{D_{z}} &\mathbb{E}_{z^{*} \sim p(z)}[logD_{z}(z^{*})] + \\
&\mathbb{E}_{x \sim p_{data}(x)}[log(1-D_{z}(E(x)))].
\end{aligned}
\end{equation}
Besides, $D_{img}$ forces $G$ to generate photo-realistic and plausible faces for an arbitrary $h$ and $l$, which can be trained along with $G$ by a similar token with Eq. (\ref{eq:dz}). Specifically,
\begin{equation}
\begin{aligned}
\mathop{\min}\limits_{G} \mathop{\max}\limits_{D_{img}} &\mathbb{E}_{x,l \sim p_{data}(x,l)}[logD_{img}(x,l)] + \\
&\mathbb{E}_{x,l \sim p_{data}(x,l)}[log(1-D_{img}(G(E(x),l)))].
\end{aligned}
\end{equation}







Finally the objective function becomes
\begin{equation}
\begin{aligned}
\mathop{\min}\limits_{E,G} &\mathop{\max}\limits_{D_{z},D_{img}} L(x,G(E(x),l)) \\
&+ \mathbb{E}_{z^{*} \sim p(z)}[log D_{z}(z^{*})] \\
&+ \mathbb{E}_{x \sim p_{data}(x)}[log(1-D_{z}(E(x)))] \\
&+ \mathbb{E}_{x,l \sim p_{data}(x,l)}[logD_{img}(x,l)] \\
&+ \mathbb{E}_{x,l \sim p_{data}(x,l)}[log(1-D_{img}(G(E(x),l)))]. \\
\end{aligned}
\end{equation}

\subsection{Genetic Mappings via DNA-Net}

We propose DNA-Net to map face features of a pair of parents to a child (see Figure~\ref{fig:dnanet}). As mentioned, DNA-Net is made-up of two networks, \ie a feature-to-genes encoder network $T_{fg}$ and a genes-to-feature decoder network $T_{gf}$. During the encoding process, given an input face feature vector $h \in R^{n}$, $T_{fg}$ produces a gene vector $g \in R^{m}$, where $n$ and $m$ are dimensions of the feature vector and gene vector respectively. During the decoding process, given a gene vector $g \in R^{m}$, the decoder $T_{gf}$ will output a feature vector $h \in R^{n}$. For the complete generation process, $T_{fg}$ predicts the gene vectors for both parents, while $T_{gf}$ maps the genes-to-features for the child.

When the gene vectors of the parents are obtained from $T_{fg}$, there are two ways to implement random selection process in Eq.~(\ref{eq:gene1}). Since the convergence of a neural network requires a certain structure, the randomness in $S(\cdot)$ should eliminate. This can be done in two ways: (1) use a determined random seed when training; (2) use a determined rule to select which parent will pass down which gene elements to child (\ie the parent for which particular genes of the child are inherited). We follow (2) in this work. Specifically, our selection rule keeps the genes with maximum values of the two parents. During testing, along with the selection rule, the DNA-Net can also use a random 0-1 sequence for genes selection from parents to children to generate additional children (\ie siblings).

The training process of DNA-Net is as follows. Given a triplet set of family images $(x_{f},x_{m},x_{c})$, we first extract facial features $(h_{f},h_{m},h_{c})$ from the trained encoder $E$ in Eq. (\ref{eq:feat1}). They are then used as the inputs and ground truth of DNA-Net. The objective of DNA-Net is to generate similar features as $h_{c}$. Therefore, the loss over the triplet set is defined as
\begin{equation}
\begin{aligned}
\mathop{\min} \limits_{T_{fg},T_{gf}} ||T_{gf}(S((T_{fg}(h_f),T_{fg}(h_m)))-h_{c}||_{2}
\end{aligned}
\end{equation}

Due to the uniform distribution constraint on $h$ in \gls{caae}, the output of DNA-Net $h_c$ should also follow the same distribution. So, a discriminator $D_h$ is trained along with $T_{fg}$ and $T_{gf}$. The loss that regularizes DNA-Net's output is defined as
\begin{equation}
\begin{aligned}
\mathop{\min} \limits_{T_{fg},T_{gf}} &\mathop{\max}\limits_{D_{h}} \mathbb{E}_{z^{*} \sim p(z)}[logD_{h}(z^{*})]+ \\
&\mathbb{E}_{h_{c}\sim T(h_{f},h_{m})}[log(1-D_{h}(T(h_{f},h_{m}))],
\end{aligned}
\label{eq:dnaminmax}
\end{equation}
 $T(h_{f},h_{m})=T_{gf}(S((T_{fg}(h_f),T_{fg}(h_m)))$ is the output.

\begin{figure}[t!]
    \centering
    \begin{subfigure}[t]{0.5\textwidth}
        \centering
        \includegraphics[width=\textwidth]{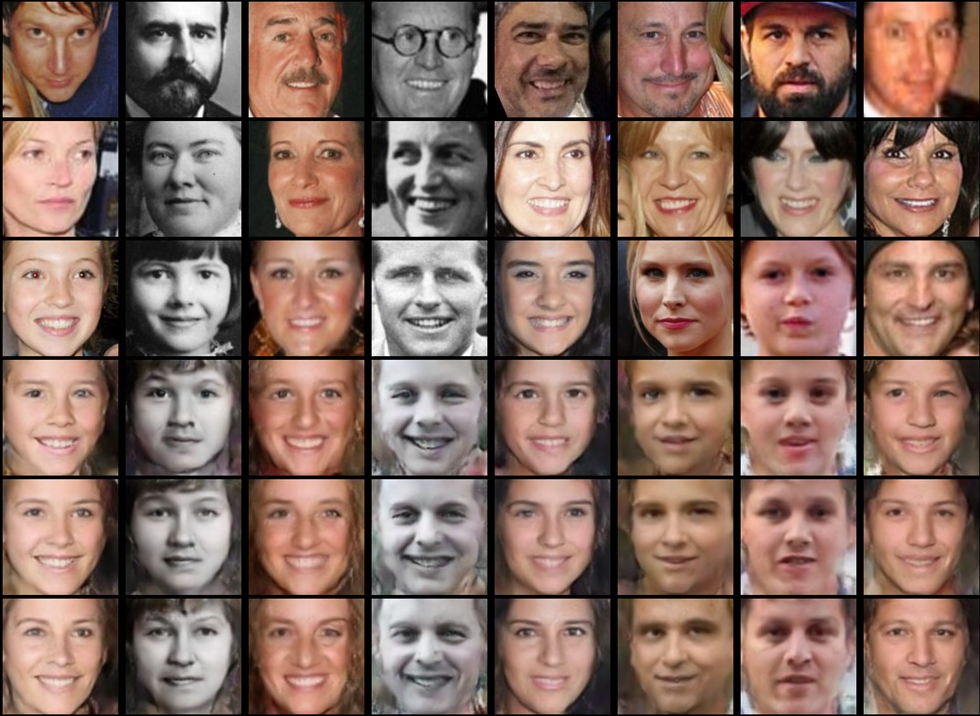}
        \caption{Across ages (\ie 10, 20, 30 years old from row 4-6, respectfully).}
        \label{fig:montage:age}
    \end{subfigure}%
    \\
    \begin{subfigure}[t]{0.5\textwidth}
        \centering
        \includegraphics[width=\textwidth]{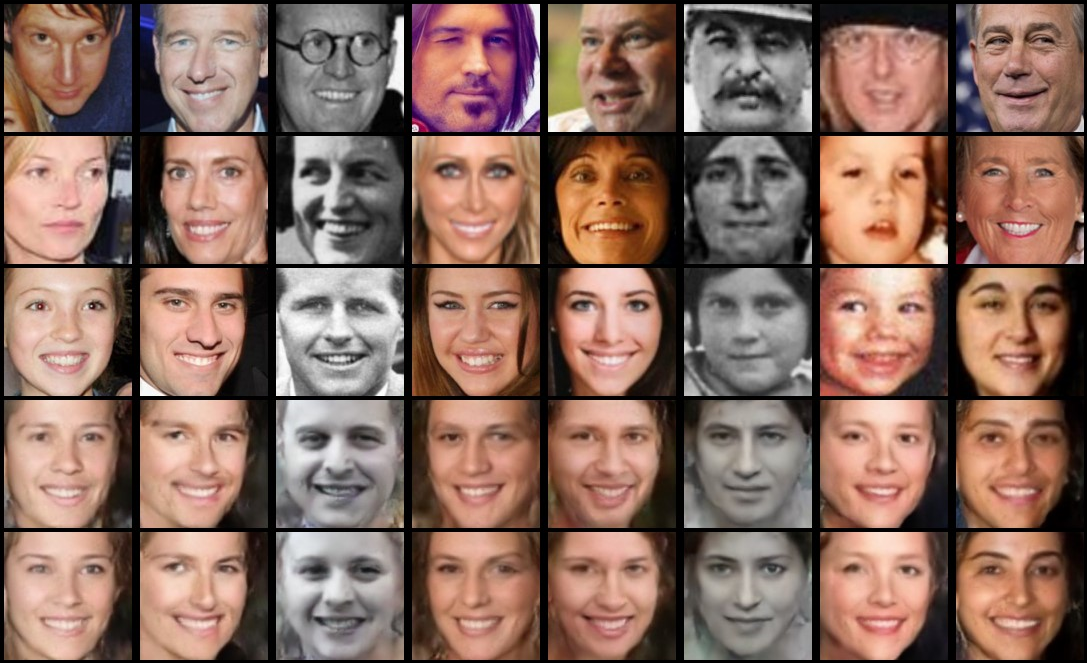}
        \caption{Across gender (\ie male-to-female from row 4-5, respectfully).}
         \label{fig:montage:gender}
    \end{subfigure}
    \caption{First three rows are real families face images which are similar to Figure~\ref{fig:samles results}. The last three and two rows are generated face images with different ages (a) and gender (b).}
\end{figure}

\section{EXPERIMENTS}\label{sec:typestyle}
This section first introduces the data, and then details the implementation. Also, our model is evaluated qualitatively and quantitatively in several experiments, specifically, conditional face generation, kinship verification, human evaluation, and heritable mappings.
\subsection{Datasets}

\textbf{UTKFace}~\cite{zhang2017age} is used to train \gls{caae} model, which divides the images into 10 age groups (\ie 0-5, 6-10, 11-15, 16-20, 21-30, 31-40, 41-50, 51-60, 61-70, and 71-80 years old). For this, a 10-dim one-hot vector is used to represent the age. For the gender, another 10-dim one-hot vector is formulated. UTKFace datasets  is a large-scale face dataset with wide age span (ranging from 0 to 116 years old), containing over 20,000 aligned and cropped face images with labels for age and gender.

\textbf{FIW}~\cite{robinson2018visual, robinson2016families} contains 1,000 families, over 11,000 persons, and is the largest kinship recognition dataset up to date. This gave us 1,997 father-mother-child face sets selected at random, with 1,600 used for training and the remaining 397 for testing.

\subsection{Implementation}
The implementation of \gls{caae} is the same as \cite{zhang2017age}. With the \gls{caae} model trained, the feature vectors $h_x$ of all faces in the father-mother-child sets could be generated and used to train the DNA-Net. In our experiment, dimensions of the feature vectors $h_x\in R^n$ and genes vectors $g_x\in R^m$ are both set to $n=m=100$. In DNA-Net, $T_{fg}$ and $T_{gf}$ are both 3-layer fully connected networks. \gls{caae} and DNA-Net were optimized using Adam optimizer \cite{Kingma2015AdamAM} with a learning rate of 0.0001.

\begin{figure}[t!]
    \centerline{
        \includegraphics[width=\linewidth, trim={0 0 0 1.7cm},clip]{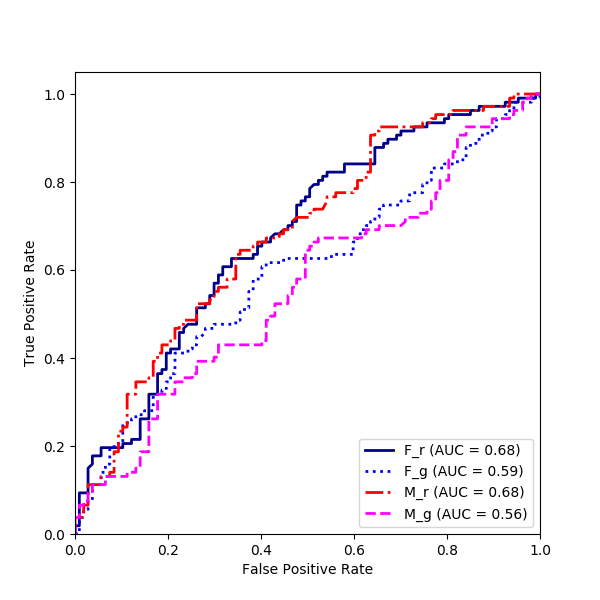}}
    \caption{ROC curve for verification evaluation. Legend items translate to father (F) or mother (M) and real (r) or generated (g) children.}
    \label{fig:roc}
\end{figure}

\subsection{Face Generation in Multiple Conditions}\label{subsec:facegen}
Figure~\ref{fig:samles results} shows examples of generated children's face images. As can be seen, the generated images have a high visual quality and clearly resemble one of the parents. For example, the mouth or eyes of generated children's face look like either their father (\eg fourth column) or mother (\eg second column). All results are with high quality, indicating that DNA-Net learns a mapping from feature space.

Benefiting from the novel two-stage generation process, our model can generate children faces at different ages and genders by changing input of age and gender labels. Samples of children with different ages are shown in Figure~\ref{fig:montage:age}, and those in different genders are shown in Figure~\ref{fig:montage:gender}. Clearly, we can observe the aging progress from juvenile to young people to middle-age in row 4-6 (\eg second column, Figure~\ref{fig:montage:age}).

\begin{figure}[t!]
\centerline{
\includegraphics[width=\linewidth]{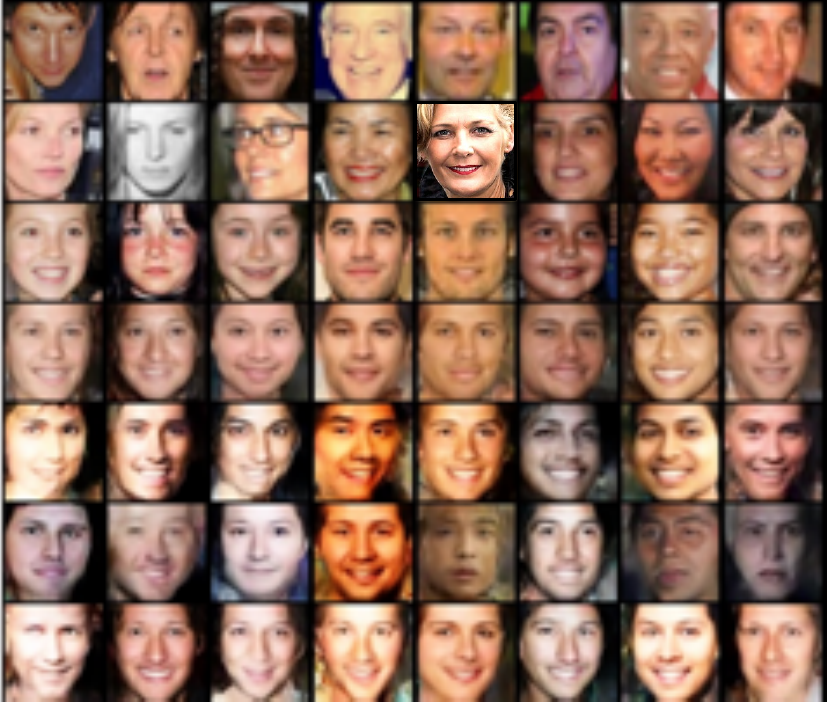}}
\caption{Samples of sibling generation. First three rows are real face images of families like in Figure~\ref{fig:samles results}. The last four rows are child faces generated with different random seeds.}
\label{fig:res}
\end{figure}

We also generate sibling faces of the child by using sequences from the random selection process instead of the determined rule for training. Some generated sibling faces are shown in Figure~\ref{fig:res}.


\begin{figure}[t!]
\centerline{
\includegraphics[width=\linewidth]{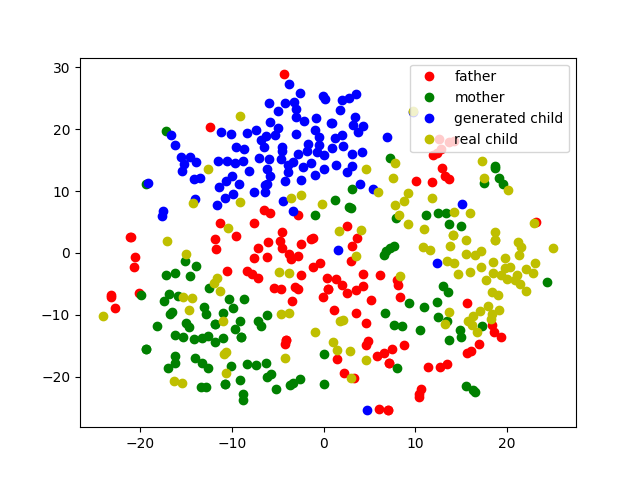}}
\caption{Visualization of facial feature distribution of fathers, mothers, children, and generated ones. Red points represent the feature of fathers, green for mothers, yellow for real children, blue for generated children, respectively. Best viewed in color.}
\label{fig:distribution}
\end{figure}

\subsection{Quantitative evaluation}\label{subsec:quantifyeval}
To quantify the performance of the proposed, we evaluated via kinship verification. Both learning model (CNN) and human subject performance are evaluated. Experimental settings and results are described in the following subsections and shown in Fig~\ref{fig:verresults}.



\subsubsection{Kinship Verification}

To evaluate the quality of generated faces, we used a pre-trained  kinship verification classifier to identify whether the generated child's image can be classified as the child of a given parent. The more generated images that can fool the classifier, the better the performance of the generation method. In this paper, the pre-trained kinship verification classifier is a FaceNet network fine-tuned on FIW~\cite{robinson2018visual}. 

We randomly sampled 100 families, with each consisting of a mother, father, and child. For each set of parents, a child's face was generated using our model. We then evaluated kinship verification accuracy on both the real and generated face images, with another 100 negative samples added to the test set. Thus, the same number of negatives was used for both the real and the generated cases. The generated children faces scored a verification accuracy of 58.89\% (with father) and 57.01\% (with mother), while the real children achieved 67.29\% and 73.83\%, respectively. Figure~\ref{fig:roc} shows the ROC curves for each cases.

To measure the identity similarity between the real child images and the generated ones, we use pre-trained FaceNet model to extract identity features from both faces, where the training data are totally independent from FIW. Then, a similarity score is computed between every two extracted features using cosine distance. The average distance of 100 real-to-generated pairs is 0.90, compared to 0.94 for generated faces and random real faces. This means the generated faces are a little closer to the real ones. In addition, we visualize the low-dimensional distribution of facial features from generated faces, real ones, and parents respectively by t-SNE \cite{maaten2008visualizing}. Figure~\ref{fig:distribution} shows the distribution of the face features-- those of generated children are more clustered, and with small overlap with real ones. This may be due to lack of large training images and complex genetic mechanism. However, we can see that the feature distributions of the generated child face is as close to the faces of the parents as it is to the faces of the real child. This is consistent with the verification results.

\subsubsection{Human Evaluation}\label{subsec:humaneval}
We asked human participants to vote on child images (real or generated) that were thought to be the true child of a pair of parents. In other words, we randomly selected 30 parent pairs from the verification set. Thus, facial images of each parent pair were shown next to their actual alongside the generated (order of the actual and generated faces were set at random, while the father was proceeded by the mother on the left side). The task was to determine the child that was the descendent of the parent pair. In other words, the volunteers picked the face of the child that resembled the parents more. Hence, each pair included a generated face of the child. We created a Google Form to distribute, and used university email lists and social media for recruiting volunteers. In total, 35 volunteers partook. Note that no volunteers had prior knowledge that some of the faces were generated (\ie we just asked which child is the true descendant).

\begin{figure}[!t]
    \begin{subfigure}[t]{0.5\linewidth}
        \centering
        \vspace{-1.25in}
        \begin{tabular}{rcc}
            \footnotesize \textbf{Pair-Type} &   \footnotesize \textbf{CNN} & \footnotesize \textbf{Human} \\
            \midrule
            \footnotesize Father-Real  &\footnotesize  67.29  &\footnotesize 38.88\\
            \footnotesize Father-Gene.  &\footnotesize  58.89  &\footnotesize 61.12\\
            \midrule
            \footnotesize Mother-Real   &\footnotesize  73.83  &\footnotesize 40.55\\
            \footnotesize Mother-Gene.  &\footnotesize  57.01  &\footnotesize 59.45\\
            \midrule
            \footnotesize Avg.-Real &\footnotesize  70.56 &\footnotesize 39.71\\
            \footnotesize Avg.-Gene. &\footnotesize  57.95 &\footnotesize 60.29\\
            \midrule
        \end{tabular}
        \subcaption{}
        \label{fig:verscores}

    \end{subfigure}
    \quad
    \centering
    \begin{subfigure}[t]{0.41\linewidth}
        \centering
        \includegraphics[width=3.0cm]{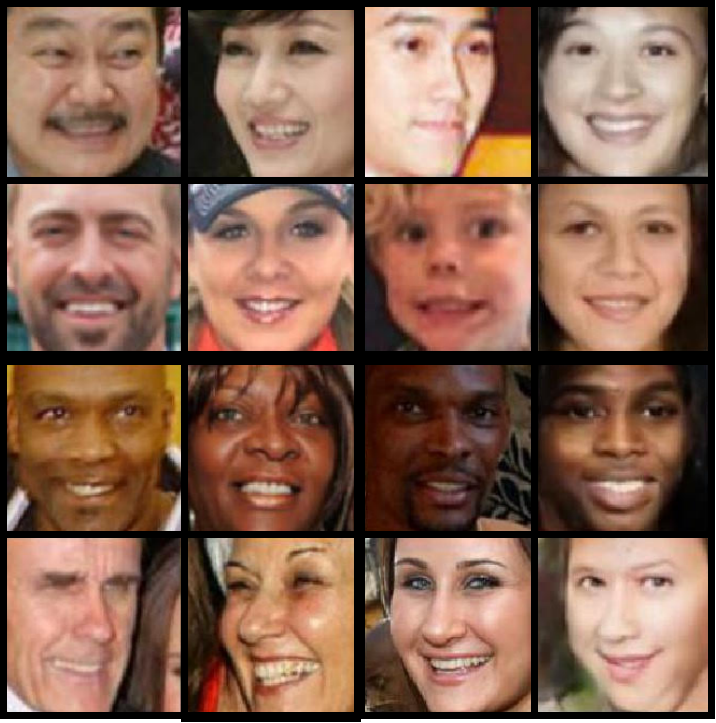}
        \subcaption{}
        \label{fig:verfaces}
    \end{subfigure}
    \caption{Kinship verification scores (\%) for real and generated children (a). Face samples shown are those that CNN and most humans agree (b): parents (columns 1-2 are father and mother, respectfully) and children (columns 3-4 are actual and generated child, respectfully). Top 3 rows are samples of generated children scored highest and accumulated most votes. To the contrary, the bottom row are real children that scored highest and received most votes. }
    \label{fig:verresults}
\end{figure}

The generated children obtained more votes than the actual. Specifically, about 60.29\% of the generated stumped the user into believing it was the true child, which was measured by the number of votes. Thus, the faces generated by the proposed appeared more genuine than that of the actual child to humans (see Figure  ~\ref{fig:verfaces}).

\subsubsection{Heritability Maps}
\begin{figure}[t!]
\centering
\begin{subfigure}[t]{0.45\linewidth}
\centering
\includegraphics[width=3.5cm]{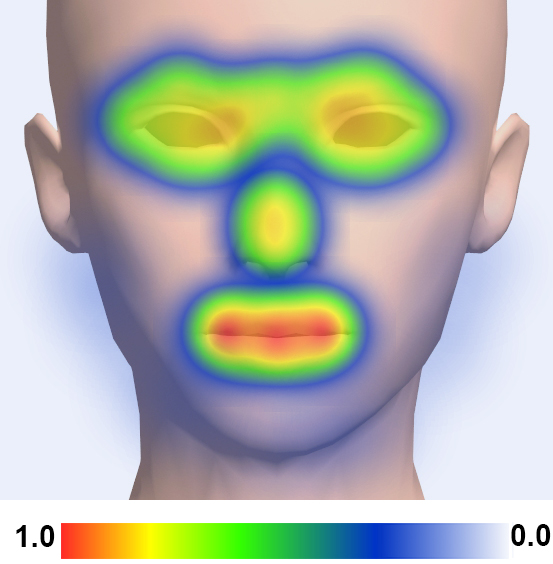}
\subcaption{Generated child}
\end{subfigure}
\quad
\begin{subfigure}[t]{0.45\linewidth}
\centering
\includegraphics[width=3.5cm]{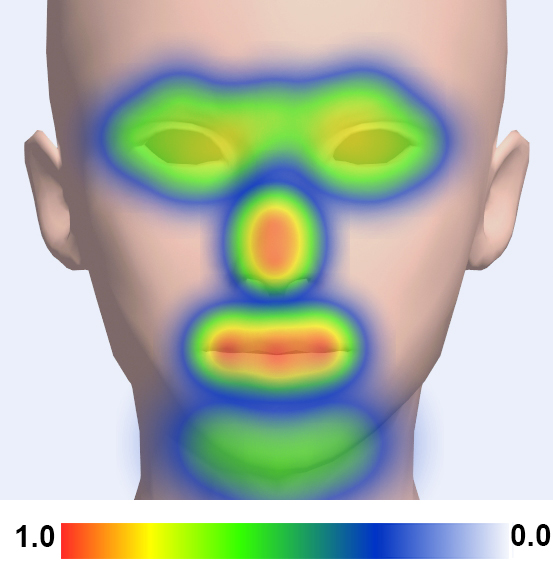}
\subcaption{Real child}
\end{subfigure}
\caption{Heritability map represents the estimated salience about facial landmarks. Best viewed in color.}
\label{fig:heritablity}
\end{figure}

It is evident that the human face consists of complex traits under strong genetic control. To further explore heritability of facial traits, we study the geometric similarity of face image pairs. Here, we compare the shape features of four parts of face, \ie eyes, nose, mouth, and chin, between parents and child. In detail, we select 20 pairs of front faces of real child and parents, generated child and parents, respectively, from above testing images. We detect the landmarks of faces and connect them into lines. After that, Hu invariant moment (\cite{hu1962visual}) is computed to represent the shapes of the four facial parts. Accumulative cosine distances are then utilized to represent heritablility. Figure~\ref{fig:heritablity}a shows the heritability map of generated child face. It can be seen that mouth region has high similarity with parents. For the real child face (see Figure~\ref{fig:heritablity}b), like the mouth, the nose region is highly similar. Besides, the chin regions are potential evidence for genetics. These results are consistent with findings in genetics \cite{debruine2009kin}.

\section{Conclusion}
In this paper, we investigate a multidisciplinary problem of children face generation from their parents which resides in the intersection of computer vision, biology and genetics. We hope to open a gate for visual face modeling for genetic combination and expression. To this end, we propose a novel DNA-Net to construct the transformation and random selection process from parents' genes to child's ones. Furthermore, our model could generate face images of children of different ages and genders by the leverage of CAAE model. Quantitative and qualitative experimental results show the generated children faces have high similarity with parents as well as similar heritability with real children. Our study could be useful in a varity of applications, ranging from population genetics and gene-mapping studies, to face modeling and reconstruction applications.

{\small
\bibliographystyle{unsrt}
\bibliography{egbib}
}

\end{document}